# Leveraging Computerized Adaptive Testing for Cost-effective Evaluation of Large Language Models in Medical Benchmarking

**Tianpeng Zheng [1,2*], Zhehan Jiang [1,3*], Jiayi Liu [4], Shicong Feng [4]**

[1] Institute of Medical Education, Health Science Center, Peking University, Beijing, China

[2] School of Public Health, Peking University, Haidian District, Beijing, China

[3] Peking University Health Science Center-Chaoxing Joint Laboratory for Digital and Smart Medical, Peking University, Beijing, China

[4] Graduate School of Education, Peking University, Haidian District, Beijing, China

**Corresponding author**：Zhehan Jiang

**E-mail address:** jiangzhehan@bjmu.edu.cn

[*] Co-first authors; Tianpeng ZHENG and Zhehan JIANG contributed equally to this manuscript

# Leveraging Computerized Adaptive Testing for Cost-effective Evaluation of Large Language Models in Medical Benchmarking

## Abstract

The rapid proliferation of large language models (LLMs) in healthcare creates an urgent need for scalable and psychometrically sound evaluation methods. Conventional static benchmarks are costly to administer repeatedly, vulnerable to data contamination, and lack calibrated measurement properties for fine-grained performance tracking. We propose and validate a computerized adaptive testing (CAT) framework grounded in item response theory (IRT) for efficient assessment of standardized medical knowledge in LLMs. The study comprises a two-phase design: a Monte Carlo simulation to identify optimal CAT configurations and an empirical evaluation of 38 LLMs using a human-calibrated medical item bank. Each model completed both the full item bank and an adaptive test that dynamically selected items based on real-time ability estimates and terminated upon reaching a predefined reliability threshold (standard error ≤ 0.3). Results show that CAT-derived proficiency estimates achieved a near-perfect correlation with full-bank estimates ($r$ = 0.988) while using only 1.3 percent of the items. Evaluation time was reduced from several hours to minutes per model, with substantial reductions in token usage and computational cost, while preserving inter-model performance rankings. This work establishes a psychometric framework for rapid, low-cost benchmarking of foundational medical knowledge in LLMs. The proposed adaptive methodology is intended as a standardized pre-screening and continuous monitoring tool and is not a substitute for real-world clinical validation or safety-oriented prospective studies.

## 1. Introduction

The integration of large language models (LLMs) into healthcare—ranging from clinical documentation to decision support—holds promise for improving diagnostic accuracy, workflow efficiency, and access to medical expertise[1]. Yet the responsible deployment of such systems hinges on reliable, scalable methods to evaluate their foundational knowledge. Current practice relies heavily on static, fixed-length benchmarks such as, MedQA[2], MMLU-Med[3], and MedBench[4]. While useful for initial comparisons, these benchmarks are poorly suited for ongoing model monitoring in a rapidly evolving technological landscape[5].

Three interrelated limitations undermine their utility as routine evaluation tools. First, they are computationally and financially expensive[6–8]. Large-scale medical benchmarks often comprise tens of thousands of multiple-choice questions (e.g., MedQA: 12,723 items[2]; MedMCQA: 6,150 items[9]; MedBench: 40,041 items[4]; Med-HALT: 59,254 items[10]), leading to substantial token consumption and API costs—particularly for proprietary models. At current pricing, a single full-benchmark evaluation can exceed $1,000. This becomes particularly problematic given that LLMs' medical performance can improve by nearly 20% within a single year[11], necessitating timely and repeated evaluation to accurately track progress—yet the substantial financial and operational burden makes such monitoring difficult to sustain. Second, most widely used benchmarks consist of publicly available items, raising serious concerns about data contamination: if test questions have been included in a model's training corpus, performance reflects memorization rather than genuine knowledge or reasoning[12,13]. Third, conventional benchmarks yield only coarse-grained metrics (e.g.,

overall accuracy), which lack the statistical precision needed to differentiate among high-performing models or detect subtle performance regressions over time[12].

Psychometrics theory offers a principled alternative. Item Response Theory (IRT) models the probability of a correct response as a function of an examinee's latent proficiency ($\theta$) and item characteristics (e.g., difficulty $\beta$, discrimination $\alpha$), calibrated on representative human samples. Once calibrated, items form a stable measurement scale, enabling comparable scoring across different subsets—a property essential for adaptive testing. In this work, we adopt an operational interpretation: $\theta$ serves as a standardized index of performance on a specific, secure knowledge bank under fixed conditions, without implying equivalence to human cognition[14].

Computerized Adaptive Testing (CAT), built upon IRT, is the gold standard in high-stakes human assessments (e.g., nursing and medical licensure exams). By selecting items that maximize information about the current ability estimate, CAT achieves equivalent precision with far fewer questions[15]. Although LLMs differ fundamentally from human test-takers, recent studies suggest their response patterns on structured knowledge tasks exhibit sufficient consistency to support IRT-based modeling[16]. Another study found that item parameters estimated from LLM responses highly correlated with those from expert predictions (Spearman's $\rho > 0.8$) [17]. Moreover, analyses of ChatGPT show structured, non-random response patterns consistent with trait-like reasoning abilities[18].

To overcome the aforementioned risks of static benchmarks, we introduce a novel framework for CAT grounded in IRT to evaluate standardized medical knowledge in LLMs. This approach leverages a secure, non-public national medical item bank calibrated on real human examinees to eliminate data contamination while enabling psychometrically rigorous measurement—a necessary prerequisite for safe clinical AI deployment.

We pursue a two-phase investigation: (1) Monte Carlo simulations to compare the adaptive item selection strategy against a random selection baseline and characterize trade-offs between test length and measurement precision under different stopping rules; (2) an empirical evaluation of 38 LLMs. This work establishes the first empirical validation of CAT for medical LLM evaluation, demonstrating that: A psychometrically sound assessment protocol can be achieved with drastically reduced item count while preserving measurement validity; Cost and time barriers to routine benchmarking can be overcome without compromising comparative precision. To corroborate the construct validity of our CAT-derived proficiency estimates—and to rule out that performance reflects MCQ-specific heuristics rather than genuine medical knowledge—we further validated our framework against an independent benchmark called LLMEval-Med Medical Knowledge subset[19] comprising real-world clinical scenarios.

Critically, our framework does not replace clinical validation—it enables it by providing a scalable, protocol-controlled foundation for pre-deployment screening and continuous monitoring. By transforming LLM evaluation from a resource-intensive bottleneck into a routine operational capability, this approach resolves a critical barrier to the responsible development and oversight of medical LLMs.

## 2. Methodology

### *2.1 Overview of the Psychometric Framework and Study Design*

To address the critical inefficiencies of static benchmark evaluations for LLMs, we designed a two-phase study. Phase 1 employed Monte Carlo simulations to optimize CAT configurations in silico. Phase 2 empirically validated the selected protocol across a diverse cohort of 38 LLMs under standardized conditions.

### *2.2 Theoretical Foundation: Item Response Theory*

Item Response Theory (IRT), regarded as a modern test theory, models the probability of a correct response as a function of an examinee's latent proficiency ($\theta$) and item characteristics, independent of the specific test-taker sample—unlike Classical Test Theory, which yields sample-dependent scores[20, 21]. We adopted the two-parameter logistic (2PL) model, which is widely used in well-known assessments worldwide (e.g., Trends in International Mathematics and Science Study, TIMSS[22]; Program for International Student Assessment, PISA[23]), for dichotomously scored items. Statistically, 2PL model is defined as:

$$P\left(Y_{j=1}\middle|\theta\right) = \frac{exp[\alpha_j(\theta-\beta_j)]}{1+exp[\alpha_j(\theta-\beta_j)]},$$

where $Y_j$ denotes the response to item $j$ (1 = correct, 0 = incorrect), $\alpha_j$ is the discrimination parameter of item $j$, indicating how well the item can differentiate between examinees with abilities above and below the item's difficulty. $\beta_j$ is the difficulty parameter of item $j$, representing the ability level at which an examiner has a 50% probability of answering the item correctly. IRT places item parameters and examinee ability on a common interval scale (typically *Mean* = 0, *SD* = 1). This allows for the direct comparison of abilities even when examinees respond to entirely different sets of items, forming the theoretical basis for Computerized Adaptive Testing[24].

### *2.3 The Computerized Adaptive Testing (CAT) System*

Computerized Adaptive Testing is an algorithmic application of IRT designed to estimate an examinee's ability with maximal cost-effectiveness and precision by dynamically tailoring the test to their performance level. Our CAT system implemented a cyclic process with four core components demonstrated in Figure 1:

(1) Initialization: For each test-taker (LLM), the CAT was initiated by randomly selecting a first item from the full bank, ensuring an unbiased starting point[15,25].

(2) Ability Estimation: After each response, the current ability estimate $\hat{\theta}$ was updated using the Expected A Posteriori (EAP) method with a standard normal prior, based on the pattern of all responses given so far[26]. Technical Details of the EAP method is provided in **Appendix A.**

(3) Item Selection: We implemented two distinct item selection strategies. As a baseline for comparison, we employed Random Selection (RS): Uniform sampling without replacement. For the adaptive condition, we utilized the Maximum Fisher Information (MFI): Selecting the unused item that maximizes information at the current $\hat{\theta}$, thereby minimizing the uncertainty of the estimate most effectively[27] (see **Appendix B).**

(4) Stopping Rule: The test terminated when a pre-defined criterion was met. Our study evaluated two distinct types of rules to explore the trade-offs between test length and measurement precision, thereby identifying the most cost-effective configurations. Fixed-Length Rules means that tests stopped after a pre-set number of items (e.g., 50, 100, 150, 200, 300, 500 items). On the other hand, Variable-Length/Precision-Based Rules denote that tests stopped once the standard error (SE) of the ability estimate fell below a specified threshold. Based on the IRT formula: $reliability(\theta) = 1 - SE(\theta)^2$, we investigated SE

thresholds of 0.500, 0.447, 0.387, 0.316 and 0.224, corresponding to target reliability levels of 0.750, 0.800, 0.850, 0.900 and 0.950, respectively. These levels were chosen based on previous studies[15,25].

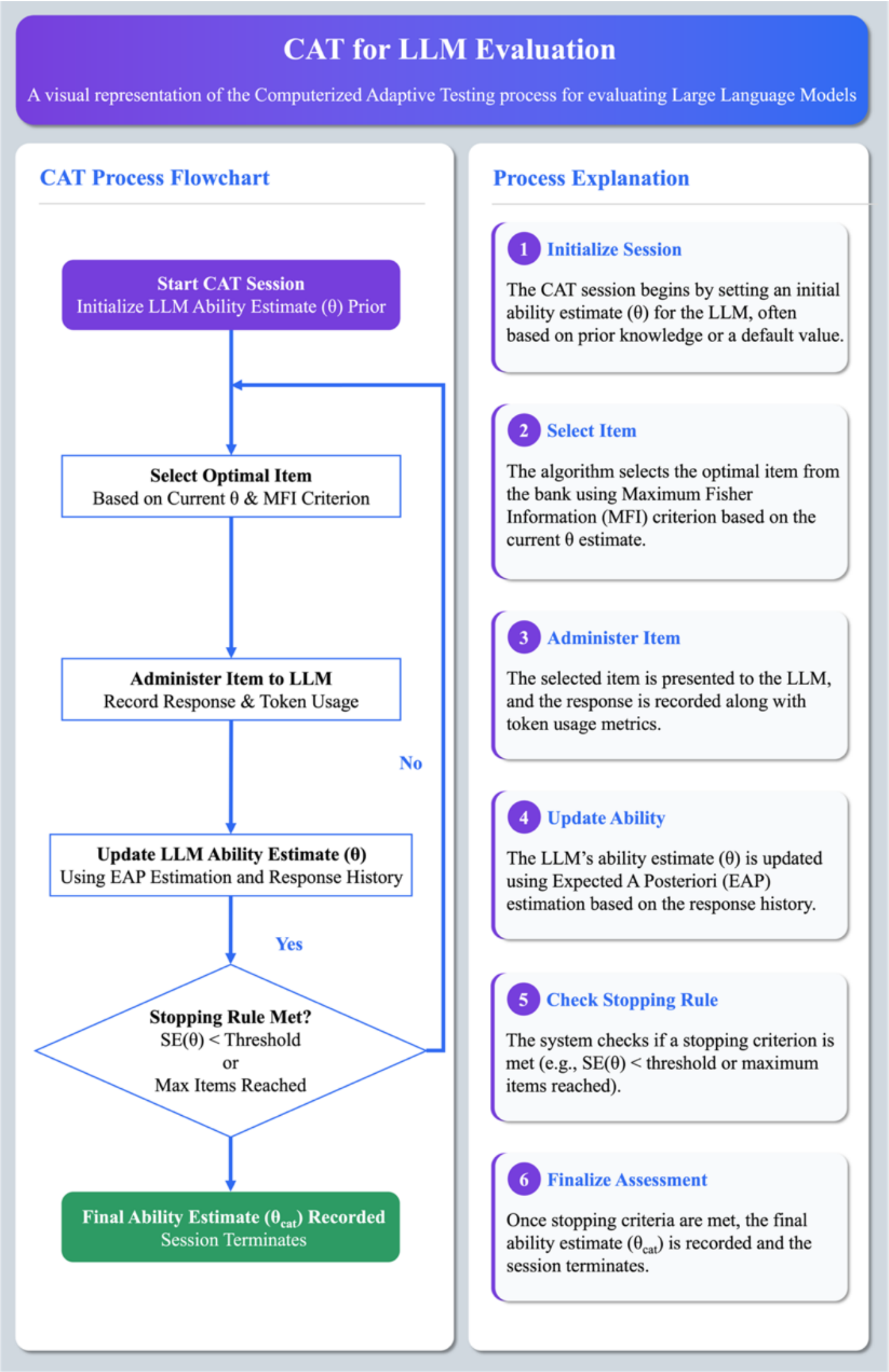

**Figure 1.** Schematic illustration of the CAT process cycle

### *2.4 Item Bank: A Secure, Human-Calibrated Foundation*

The validity of any CAT system hinges on a large, well-calibrated item bank that assesses medical knowledge. To minimize benchmark leakage risk and benchmark contamination issues[12,13], we utilized a secure, non-public item bank sourced from China's

National Center for Health Professions Education Development, used for the Competency Test for Clinical Medicine Undergraduates (CTCMU), of which item bank consists of 2,815 high-quality multiple-choice questions (MCQs) across multiple subdomains, with item parameters calibrated using responses from 40,000+ examinees at their fourth or fifth year of medical school[28].

Adopting MCQs in this study warrants explicit justification. While open-ended or free-response formats may better capture certain aspects of clinical reasoning, our use of MCQs is grounded in four interrelated practical and psychometric considerations: (1) the availability of a large-scale, nationally administered medical examination with rigorous human calibration; (2) the superior suitability of MCQs for precise item parameter estimation under IRT; (3) their established dominance in high-stakes medical assessments worldwide; and (4) empirical evidence that well-constructed MCQs can validly assess higher-order cognitive competencies. Crucially, our adaptive framework is format-agnostic in principle—it requires only calibrated items on a common $\theta$ scale. MCQs were selected not as an ideal endpoint, but as the only currently available format that simultaneously satisfies scalability, psychometric rigor, and contamination resistance at the national level. To support this rationale, we summarize key evidence in Table 1.

**Table 1**. Rationale for Using MCQs as the Primary CAT Benchmark

| Methodological Argument | Supporting Evidence |
|---|---|
| Authoritative source with human calibration | To mitigate benchmark leakage and contamination risks, we employed a secure, non-public item bank sourced from China's National Center for Health Professions Education Development. The bank contains 2,815 high-quality MCQs spanning multiple medical subdomains, with IRT-calibrated item parameters based on |
| | responses from 40,000+ senior medical students (4th/5th year)—providing the stable discrimination (α) and difficulty (β) estimates required for CAT. |
| Psychometric suitability for adaptive testing | CAT depends on IRT-derived item information and requires reliable item parameter estimates. Dichotomously scored MCQs support objective scoring and statistically stable parameter estimation relative to rubric-scored open-ended responses, where scoring subjectivity and rater effects can introduce additional noise and threaten parameter stability. This reliability is critical for maintaining the validity of adaptive item selection and ability estimation. |
| Global standard in high-stakes medical assessment | All major licensing exams use MCQs as their primary format: USMLE Step 1/2CK (USA), PLAB 1 (UK), MCCQE (Canada), AMC CAT (Australia), and China's National Competency Test for Clinical Medicine Undergraduates. Leading medical LLM benchmarks also rely on MCQs: MedQA[2], MMLU-Med[3], MedMCQA[9], and MedBench[4]. |
| Capability to assess higher-order cognition when well-designed | Well-designed MCQs can validly assess higher-order clinical reasoning[29]. Prior research indicates that when responding to high-quality clinical vignette–based MCQs, examinees engage cognitive processes equivalent to those involved in authentic clinical reasoning[30, 31]. |
| Framework is extensible beyond MCQs | Our CAT protocol depends only on calibrated item parameters—not response format. In principle, it could be applied to any scored item type (e.g., short answer, script concordance) once a sufficiently large, human-calibrated bank becomes available. We demonstrate this generality via construct validity testing on LLMEval-Med's clinical scenario subset (Appendix E). |

We evaluated the appropriateness of key IRT assumptions on the human calibration dataset. Exploratory factor analysis (EFA) supported essential unidimensionality: the first and second eigenvalues were 25.30 and 3.35, respectively (ratio = 7.55), indicating a dominant general factor underlying item responses. Local independence was examined via residual correlations and LD statistics. No item pairs exhibited residual correlations exceeding |0.20|, and no item pairs showed LD statistics above |5|, suggesting minimal evidence of local dependence and that any potential inflation of reliability due to residual item dependencies is likely negligible. This large-scale human calibration provided stable, highly reliable item parameters (i.e., $\alpha$ and $\beta$ from IRT 2PL model), creating a solid psychometric foundation. Specifically, difficulty parameters were well-centered around the mean ability of the human reference cohort (*M* = -0.01, *SD* = 0.20), spanning a range from -1.11 to 1.44. This supports precise measurement around the ability range covered by the item bank. Discrimination parameters indicated high item quality (*M* = 1.01, *SD* = 0.08), with values ranging from 0.44 to 1.52.

These assumption checks pertain to the human calibration data that define the item bank parameters and the $\theta$ scale. For each LLM, the $\theta$ is interpreted operationally as a proficiency index on this calibrated item bank under a standardized evaluation protocol, rather than as evidence of a human-like latent cognitive trait. Accordingly, the IRT assumptions support the validity of the calibrated item bank and $\theta$ scale for comparative benchmarking; they do not imply a human-like latent structure for LLMs.

### *2.5 Evaluation Metrics*

To comprehensively evaluate the performance of the CAT system, we employed a suite of metrics designed to quantify both measurement precision and cost-effectiveness across both study phases. All metrics reported in the Results section are defined as follows.

#### *2.5.1 Core Measurement Precision Metrics*

(1) *Bias:*

The average difference between the estimated and true ability values. An ideal system has a bias close to zero, indicating no systematic over- or underestimation. It is computed as follows:

$$\text{Bias} = \frac{\sum_{i=1}^{N}(\widehat{\theta_i} - \theta_i)}{N},$$

where $\widehat{\theta_i}$ represents the estimated ability for examinee $i$, $\theta_i$ is the true ability for examinee $i$, and $N$ is the number of examinees.

(2) *Root Mean Square Error (RMSE):* The square root of the average squared difference between estimated and true ability, RMSE is a comprehensive measure of overall accuracy, penalizing large errors more heavily. It is computed as follows:

$$\text{RMSE} = \sqrt{\frac{1}{N}\sum_{i=1}^{N}\left(\widehat{\theta_i} - \theta_i\right)^2},$$

(3) *Pearson Correlation (Cor):*

The correlation coefficient between the estimated and true ability vectors, with values closer to 1.0 indicating a better precision.

***2.5.2 Cost-Effectiveness Metric***

(1) Average Test Length (ATL):

The mean number of items administered per examinee to meet the stopping rule.

***2.5.3 Trade-off Analysis Metrics (Relative to Full-Bank Assessment)***

To quantify the cost-benefit of CAT, the following relative metrics were calculated using the full-bank assessment as the benchmark.

(1) *Test Length Reduction (TLR):* This metric quantifies the percentage reduction in the number of administered items for a CAT condition relative to the full-bank assessment. It serves as a direct proxy for the gains in cost-effectiveness and the reduction in computational cost. A higher TLR value indicates greater cost-effectiveness. It is calculated as:

$$TLR = \left(1 - \frac{ATL}{TTL}\right) \times 100\%,$$

where $TTL$ is the length of the full item bank.

(2) *Bias Increase Rate (BIR):*

To assess the change in systematic error, we calculated the Bias Increase Rate. This metric quantifies the percentage reduction in the magnitude of the bias (the absolute bias) for a CAT condition relative to the full-bank baseline. A positive value indicates that the CAT condition has a smaller absolute bias (an improvement), while a negative value indicates a larger absolute bias (a worsening). Therefore, higher values are better. It is calculated as:

$$BIR = \left(1 - |\frac{Bias_{condition}}{Bias_{full-bank}}|\right) \times 100\%,$$

$Bias_{condition}$ and $Bias_{full-bank}$ be the bias for a given CAT condition and the full-bank test, respectively.

(3) *RMSE Increase Rate (RIR):*

RMSE Increase Rate quantifies the percentage reduction in the magnitude of the RMSE for a CAT condition relative to the full-bank baseline. A positive value indicates that the CAT condition has a smaller absolute bias (an improvement), while a negative value indicates a larger absolute bias (a worsening). Therefore, higher values are better. It is calculated as:

$$RIR = \left(1 - |\frac{RMSE_{condition}}{RMSE_{full-bank}}|\right) \times 100\%,$$

$RMSE_{condition}$ and $RMSE_{full-bank}$ be the RMSE for a given CAT condition and the full-bank test, respectively.

(4) *Correlation Loss Rate (CLR)*:

Correlation Loss Rate quantifies the percentage loss in correlation between estimated ability and true ability of a CAT administration compared to the full-bank test. Lower values are better. It is calculated as:

$$CLR = \left(1 - \left|\frac{Cor_{condition}}{Cor_{full-bank}}\right|\right) \times 100\%,$$

$Cor_{condition}$ and $Cor_{full-bank}$ be the Pearson Correlation for a given CAT condition and the full-bank test, respectively.

***2.6 Phase 1: Monte Carlo Simulation Design***

The objective of the simulation phase was to systematically evaluate the performance and cost-effectiveness of our CAT system across a comprehensive range of ability levels and stopping rules, without the computational expense of running LLMs.

We simulated 3,600 examinees with true ability parameters $\theta$ drawn from a uniform distribution, spanning 36 discrete ability levels from -3.5 to +3.5 at intervals of 0.2. Each ability level was replicated 100 times. This uniform sampling, as opposed to a normal distribution, provides a more stringent test of the CAT algorithm's robustness across the entire ability spectrum[15, 25, 32]. For each simulated examinee, we generated responses based on the 2PL model and the known item parameters. Each examinee was assessed under 12 stopping rule conditions (7 fixed-length, 5 variable-length) and 2 item selection methods (RS and MFI). All simulations were conducted using the *catR* package in R[33].

***2.7 Phase 2: Empirical Evaluation with LLMs***

To ensure robust generalizability of our findings, we evaluated a diverse cohort of 38 LLMs. This included both proprietary API-based models (e.g., GPT-4, Claude-4-Sonnet-Thinking) and open-source models (e.g., Qwen series and MedGemma), spanning a wide range of scales (from 0.5B to over 120B parameters) and specializations (general-purpose and medical-domain). Each model underwent two distinct evaluations: (1) Full-Bank Assessment ($\theta_{Full}$): Each model was evaluated on the entire item bank to establish a high-fidelity ground truth ability estimate via EAP estimation; (2) CAT Assessment ($\theta_{CAT}$): Based on the optimal rule identified in Phase 1, each model underwent a separate CAT administration. The CAT was configured with random item initialization.

We accessed the LLMs through their publicly available API platforms, while for open-source models, we used either API access or local deployment to generate responses. The temperature parameter was fixed at 0.0 to ensure deterministic outputs. The top-$p$ value was set to 1.0 to allow sampling from the full probability distribution without truncation. Recognizing the significant influence of prompt engineering on the outputs of generative models, we standardized the input format across all datasets. A detailed list of all evaluated models with their specifications, along with the prompt and illustrative examples of questions and corresponding LLM responses, is provided in **Appendix C and D**.

For the empirical evaluation, we focused on two primary analyses: (1) Measurement Fidelity: We assessed the agreement between the full-bank and CAT assessments by calculating: (a) Pearson correlation *Cor* between $\theta_{Full}$ and $\theta_{CAT}$; (b) Spearman's rank correlation $\rho$ between the resulting model rankings; (2) Cost-effectiveness Gains from cost-effectiveness perspective: We quantified the practical resource savings by computing the TLR, total token consumption, and total wall-clock time required for the CAT evaluation compared to the full-bank baseline.

In addition, to examine the content validity of the proposed CAT framework—namely, whether it meaningfully reflects LLMs' underlying medical knowledge rather than exploiting heuristics specific to multiple-choice formats—we conducted a supplementary validation using the Medical Knowledge subset (141 items) of the LLMEval-Med benchmark[19]. This subset consists of open-ended question-answering tasks derived from real-world electronic health records, requiring free-text responses rather than option selection. By evaluating the same cohort of models on this independent, open-ended benchmark, we aimed to assess the extent to which CAT-derived ability estimates correspond to performance on authentic clinical reasoning tasks beyond structured MCQs. Detailed protocols for this validation are provided in **Appendix E**.

### *2.8. Ethical Considerations*

The proprietary item bank is not publicly available due to confidentiality agreements. The data are available from the corresponding author upon reasonable request and with permission from NCHPED. This study did not involve human participants, animals, or patient data. Testing was conducted on LLMs in accordance with their respective API terms and licenses.

## 3. Results

### *3.1 Simulation Study Results*

#### *3.1.1 Performance of Stopping Rules*

Figure 2 presents the performance of all tested stopping rules and item selection methods across four key metrics: Bias, RMSE, Correlation, and ATL. The full item bank administration, comprising all 2,815 items, served as the standard for comparison. It established the benchmark for overall accuracy by achieving the lowest RMSE of 0.100 and a near-perfect correlation of 0.9995 with the true ability scores. However, its bias was not the lowest, recorded at -0.0014 (Figure 2A).

To strictly isolate the efficiency gains attributable to the adaptive algorithm, we first compared the MFI strategy against a RS baseline. The results empirically demonstrate that the reported gains stem from algorithmic adaptivity rather than mere test shortening. Under fixed-length conditions, the adaptive approach consistently minimized measurement error more effectively than RS; for instance, at a test length of 50 items, MFI achieved an RMSE of 0.3982, significantly lower than the 0.5061 observed for the RS baseline (Figure 2B). Conversely, when targeting a specific precision level (variable-length rules), the non-adaptive random strategy required a substantially larger number of items to meet the stopping criterion. Notably, to achieve a target reliability of 0.90 (SE ≤ 0.316), the MFI algorithm required an average of only 58 items, whereas RS necessitated 89 items (Figure 2D), representing a considerable loss in efficiency without adaptivity.

Having established the superiority of the MFI strategy, we further analyzed the trade-offs among its specific stopping rules to identify the optimal configuration. Against the full-bank baseline, all adaptive conditions demonstrated dramatic gains in cost-effectiveness. The Mean Test Length (Figure 2D) shows a stark contrast between the full-bank and adaptive tests. Notably, the variable-length rules were extremely efficient: the SE ≤ 0.316 condition required an average of only 58.3 items, a 97.9% reduction in test length compared to the full bank. Even the most stringent precision rule, SE ≤ 0.224 (targeting 0.95 reliability), required only 149.1 items on average. While this reduction in test length naturally incurred an increase in measurement error, with RMSE decreasing systematically from 0.7232 for the most lenient rule (SE ≤ 0.5) to 0.1417 for the longest CAT (Length_500), it did not compromise the system's ability to evaluate examinees accurately. The correlation with true ability scores remained high across all MFI conditions (Figure 2C); even the highly efficient SE ≤ 0.5 condition achieved a correlation of 0.96, offering an optimal balance between precision and cost.

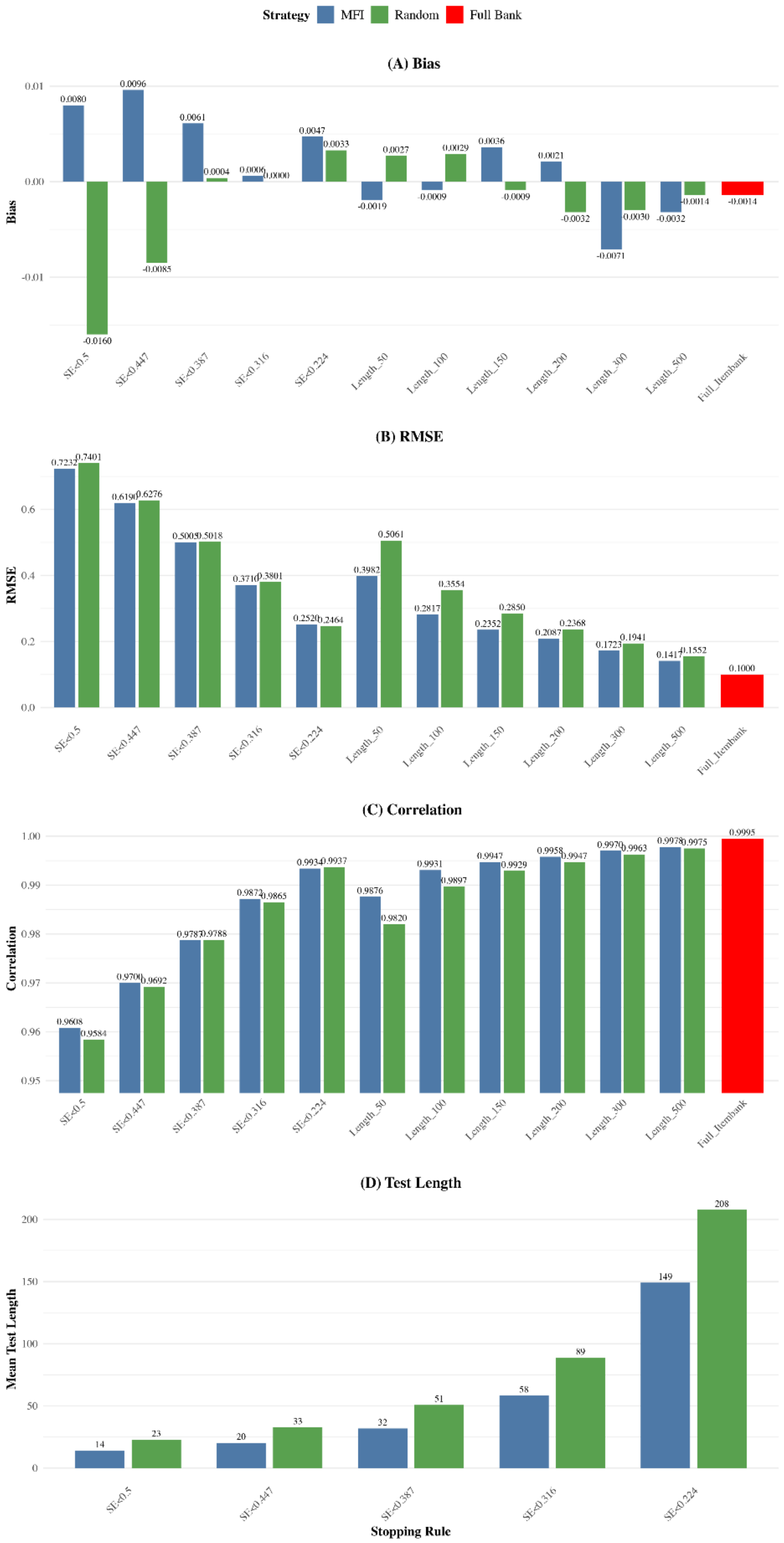

**Figure 2.** Performance Comparison of CAT from the Simulation Study

Each panel displays a key performance metric for various stopping rules and item selection methods compared to the full item bank. Metrics shown are (A) Bias, (B) Root Mean Square Error (RMSE), (C) Pearson Correlation with true ability, and (D) Mean Test Length.

### *3.1.2 Performance-Cost Trade-off Analysis*

Given the demonstrated inferiority of RS strategy across all efficiency and precision metrics in the preliminary comparison, it was excluded from this trade-off analysis. Consequently, we focused exclusively on the MFI-based configurations to identify the optimal stopping rule that balances measurement fidelity with computational cost. To visualize the trade-off between the cost-effectiveness gains and the corresponding impact on precision, we plotted the Test Length Reduction against three different precision metrics, fitting each with a LOESS smoothing curve to highlight the general trend (Figure 3). In these plots, the ideal stopping rule would be located in the top-left (e.g., Figure 3C) and top-right corner (e.g., Figure 3A/B), representing the maximum cost-effectiveness gain for the minimum precision cost (or maximum precision improvement). The distribution of the points forms a clear Pareto frontier of optimal choices.

An analysis of the trade-off with the RMSE Improvement Rate (Figure 3B) reveals a clear performance hierarchy. The SE-targeted rules consistently define the upper edge of the Pareto frontier, indicating a superior cost-effectiveness for achieving any given level of RMSE improvement. This plot features a distinct “elbow” in the region of Length_100, SE $\leq$ 0.224, and SE $\leq$ 0.316, which represents the point of diminishing returns where further gains in cost-effectiveness require a disproportionately larger sacrifice in precision.

In contrast, the trade-off with the Correlation Loss Rate (Figure 3C) reveals a different dynamic, showcasing a clear divergence between the two types of stopping rules. The fixed-length rules are clustered in the bottom-left quadrant, offering the highest fidelity (i.e., the lowest correlation loss, approaching zero) at the cost of lower cost-effectiveness.

Conversely, the SE-targeted rules are grouped in the top-right quadrant, providing maximal cost-effectiveness but incurring a slightly larger, yet still small, loss in correlation. This highlights a clear strategic choice for practitioners: fixed-length rules are optimal for minimizing correlation loss, while SE-targeted rules are optimal for maximizing cost-effectiveness.

Synthesizing the evidence from all plots, the SE ≤ 0.316 rule solidifies its position as the best overall choice. Its superiority is established on three grounds: First, it is consistently located at the “elbow” of the frontier in the clear and robust RMSE and Correlation trade-off plots, signifying a peak “return on investment”. While it does not offer the absolute lowest correlation loss, its loss is marginal (approx. 1.2%), resulting in a near-perfect correlation of 0.988. Second, it is one of the few conditions to achieve a positive Bias Improvement Rate (Figure 3A), demonstrating its unique strength in producing a more accurate, less systematically skewed estimate than even the full-bank test. Third, it achieves these precision benefits while simultaneously delivering a 97.9% reduction in test length, making it a powerful “better, faster, cheaper” solution for practical LLM evaluation.

This multi-faceted evidence strongly supports the adoption of the SE ≤ 0.316 rule for balancing the competing demands of cost-effectiveness and precision in psychometric-based LLM benchmarking.

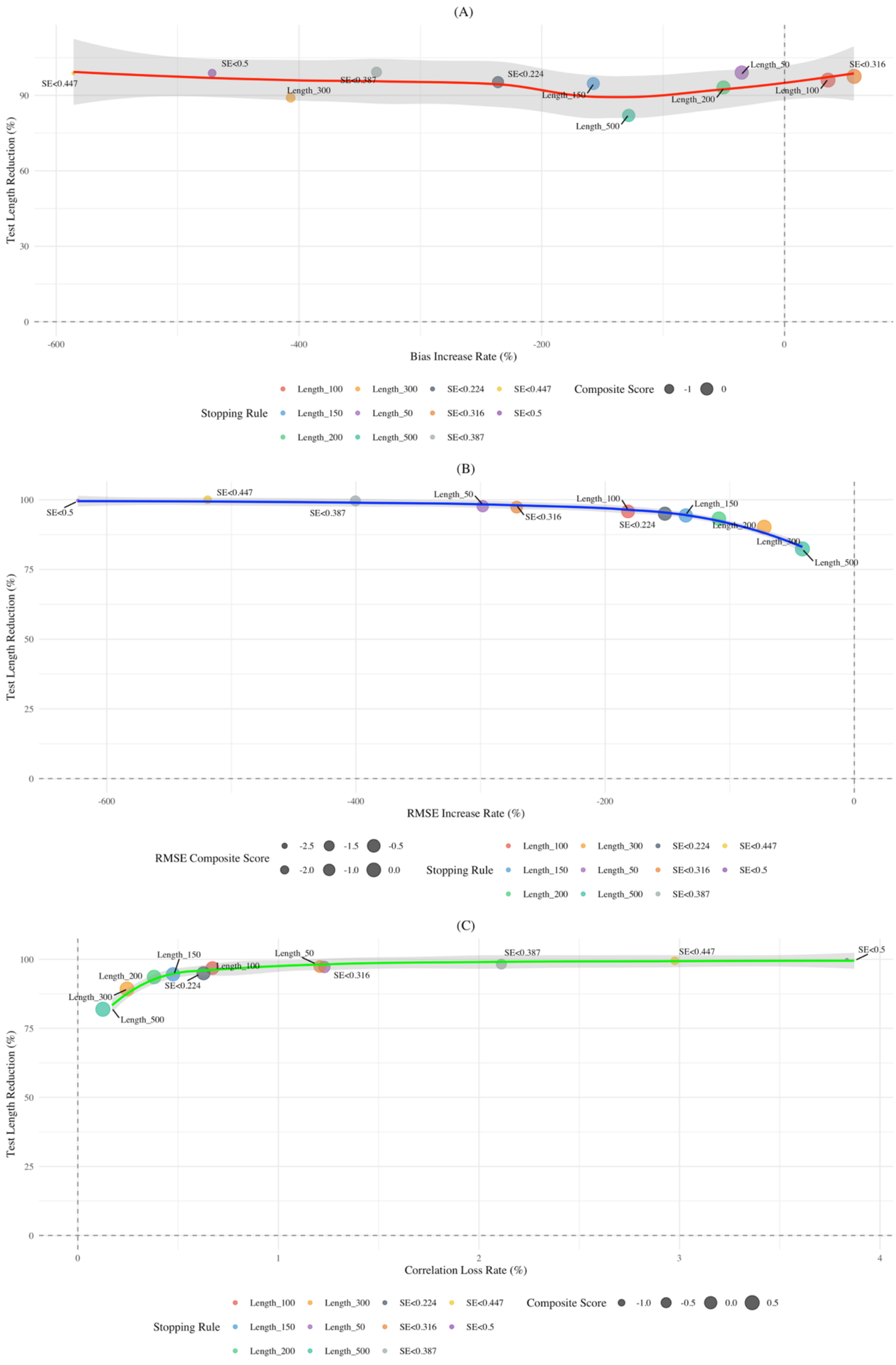

**Figure 3.** Performance-Cost Trade-off Plots

(A) Test Length Reduction vs. Bias Increase; (B) Test Length Reduction vs. RMSE Increase; (C) Test Length Reduction vs. Correlation Loss.

### *3.2 Empirical Study Results*

### *3.2.1 Validation of Measurement*

Our empirical analysis began by validating the measurement fidelity of the CAT algorithm. A detailed results for each of the 38 LLMs is provided in Table 2. Firstly, Figure 4 presents a side-by-side comparison of these $\theta_{Full}$ with $\theta_{CAT}$. The results reveal a wide spectrum of capabilities within our LLM cohort, with scores ranging from over 2.4 for state-of-the-art models down to nearly -0.9 for smaller models. The $\theta_{CAT}$ exhibited an exceptionally strong correlation with $\theta_{Full}$, achieving $r = 0.988$. As a supplementary validation, the correlation of raw accuracy scores also remained high at $r = 0.912$. This confirms that the CAT accurately captures the models' latent abilities.

More strikingly, the CAT achieved a perfect replication of the LLM leaderboard. Spearman's rank correlation coefficient, which was $\rho = 1.0$ ($p < 0.001$). Figure 5 reveals a distinct stratification of model capabilities, where the blue data points (CAT estimates) are separated by clear performance gaps. For instance, the distance between the state-of-the-art cluster ($\theta > 2.0$) and the mid-tier models is significantly larger than the measurement deviation. Consequently, the observed estimation noise was insufficient to disrupt the relative ordering of the models, resulting in a robust leaderboard replication. This result indicate that the CAT-based ranking was identical to the one produced by the exhaustive, and computationally expensive, full-bank assessment.

To corroborate the construct validity of our CAT framework and ensure that the estimated proficiency reflects genuine medical knowledge rather than MCQ-specific heuristics, we utilized a supplementary evaluation on the Medical Knowledge subset (141 items) of LLMEval-Med benchmark[19], which created from real-world electronic health records and expert-designed clinical scenarios. Detailed protocols for this validation are provided in **Appendix E**.

Due to the rapid iteration of LLMs, this analysis was performed on the 34 models that remained accessible via API as of February 2026. As shown in Supplementary Figure E1, we observed a robust linear relationship between the adaptive MCQ proficiency and generative performance. Notably, this correlation remained high and consistent across two distinct evaluators: $r$=0.829 when scored by the generalist judge (GPT-5.2), and $r$=0.831 when scored by the medical specialist judge (Baichuan-M3). This strong agreement indicates that the CAT-derived standardized knowledge proficiency index is strongly associated with performance on an open-ended medical knowledge QA benchmark.

**Table 2**. Detailed Results for Each of the 38 LLMs.

| Model | Full item bank (2815 items) | | | | CAT | | | | |
|---|---|---|---|---|---|---|---|---|---|
| | Theta | Accuracy | Total Tokens | Total Time(s) | Test Length | Theta | Accuracy | Total Tokens | Total Time(s) |
| o1 | 2.25 | 90.69% | 2262040 | 21948.43 | 73 | 2.42 | 90.41% | 75008 | 790.05 |
| gemini-pro | 2.24 | 90.66% | 4199151 | 43261.89 | 69 | 2.36 | 89.86% | 107350 | 1043.98 |
| gemini-2.5-pro | 2.17 | 90.05% | 4150473 | 41674.90 | 70 | 2.34 | 87.14% | 108122 | 1116.99 |
| deepseek-r1 | 2.00 | 88.41% | 4623837 | 127313.89 | 59 | 2.25 | 89.83% | 107159 | 3241.27 |
| claude-4-sonnet-thinking | 1.91 | 87.46% | 2012822 | 47918.66 | 54 | 2.17 | 85.19% | 39635 | 1068.06 |
| gemini-2.5-flash | 1.91 | 87.35% | 3786735 | 23407.74 | 55 | 2.09 | 89.09% | 68188 | 429.80 |
| gpt-oss-120b | 1.82 | 86.32% | 1624119 | 26510.19 | 50 | 1.97 | 80.00% | 39114 | 932.48 |
| minimax-m1 | 1.75 | 85.58% | 7131754 | 316596.65 | 48 | 1.83 | 75.00% | 133680 | 6566.40 |
| gpt-4o | 1.75 | 85.47% | 726135 | 2664.20 | 52 | 1.70 | 86.54% | 13003 | 48.47 |
| glm-4.5-air | 1.73 | 85.33% | 5246139 | 112054.24 | 49 | 1.57 | 83.67% | 94582 | 1756.69 |
| qwen3-30b-a3b | 1.56 | 82.91% | 6190960 | 48662.88 | 45 | 1.55 | 75.56% | 111964 | 886.19 |
| o1-mini | 1.53 | 82.63% | 2010210 | 11702.52 | 42 | 1.43 | 88.10% | 30748 | 158.63 |
| kimi-k2 | 1.50 | 82.24% | 777343 | 6241.36 | 41 | 1.43 | 75.61% | 10809 | 101.51 |
| gemini-2.0-flash | 1.49 | 81.99% | 755259 | 2555.54 | 41 | 1.43 | 75.61% | 10811 | 36.51 |
| gpt-oss-20b | 1.33 | 79.57% | 1868072 | 9763.62 | 37 | 1.37 | 78.38% | 27882 | 138.60 |
| gpt-4 | 1.29 | 78.72% | 734656 | 6520.99 | 38 | 1.31 | 78.95% | 9688 | 65.76 |
| glm-3-turbo | 0.99 | 73.18% | 740403 | 2842.69 | 30 | 1.03 | 73.33% | 7693 | 32.33 |
| yi-lightning | 0.98 | 72.93% | 861252 | 3906.47 | 30 | 1.00 | 73.33% | 9045 | 40.46 |
| yi-spark | 0.98 | 72.86% | 861252 | 3763.33 | 30 | 0.85 | 63.33% | 9045 | 43.31 |
| qwen2.5-72b-instruct | 0.97 | 72.82% | 756202 | 2008.90 | 32 | 0.84 | 56.25% | 8448 | 13.64 |
| qwen-max | 0.97 | 72.82% | 756063 | 1233.18 | 31 | 0.73 | 64.52% | 8076 | 22.05 |
| yi-medium | 0.93 | 72.18% | 861252 | 4138.59 | 29 | 0.66 | 68.97% | 8521 | 47.81 |
| hunyuan-standard | 0.81 | 69.95% | 775802 | 4237.40 | 28 | 0.65 | 57.14% | 7473 | 43.50 |
| qwen2.5-32b-instruct | 0.80 | 69.45% | 756063 | 1403.71 | 28 | 0.63 | 60.71% | 7321 | 13.82 |
| gpt-4o-mini | 0.74 | 68.24% | 726085 | 2151.43 | 27 | 0.58 | 66.67% | 6909 | 20.01 |
| glm-4.1v-9b-thinking | 0.74 | 68.17% | 2774268 | 21472.27 | 27 | 0.57 | 62.96% | 23438 | 178.56 |
| glm-4-air | 0.71 | 67.71% | 734857 | 3007.70 | 28 | 0.47 | 53.57% | 7257 | 36.32 |
| medgemma-27b | 0.70 | 67.28% | 784768 | 6574.93 | 26 | 0.45 | 57.69% | 7043 | 60.00 |
| qwen2.5-14b-instruct | 0.55 | 63.87% | 756063 | 1531.09 | 25 | 0.43 | 72.00% | 6706 | 13.36 |
| phi-4 | 0.54 | 63.80% | 761525 | 5351.81 | 26 | 0.42 | 65.38% | 7025 | 43.18 |
| glm-4 | 0.44 | 61.46% | 734960 | 3154.80 | 24 | 0.41 | 62.50% | 6272 | 26.29 |
| gpt-3.5-turbo | 0.25 | 56.70% | 734447 | 2188.45 | 23 | 0.36 | 56.52% | 6116 | 16.01 |
| internlm2-20b | 0.15 | 54.21% | 795604 | 6220.29 | 23 | 0.35 | 60.87% | 6568 | 50.98 |
| qwen2.5-7b-instruct | 0.10 | 53.00% | 756063 | 2178.52 | 23 | -0.10 | 43.48% | 6383 | 19.07 |
| qwen2.5-3b-instruct | -0.18 | 46.11% | 756063 | 1404.76 | 24 | -0.24 | 62.50% | 6716 | 11.44 |
| internlm2-7b | -0.20 | 45.15% | 856734 | 6762.02 | 24 | -0.34 | 37.50% | 7798 | 57.62 |
| qwen2.5-1.5b-instruct | -0.60 | 35.45% | 756063 | 1621.23 | 27 | -0.60 | 25.93% | 7623 | 15.73 |
| qwen2.5-0.5b-instruct | -0.79 | 31.55% | 756063 | 1612.30 | 29 | -0.86 | 34.48% | 8435 | 16.74 |

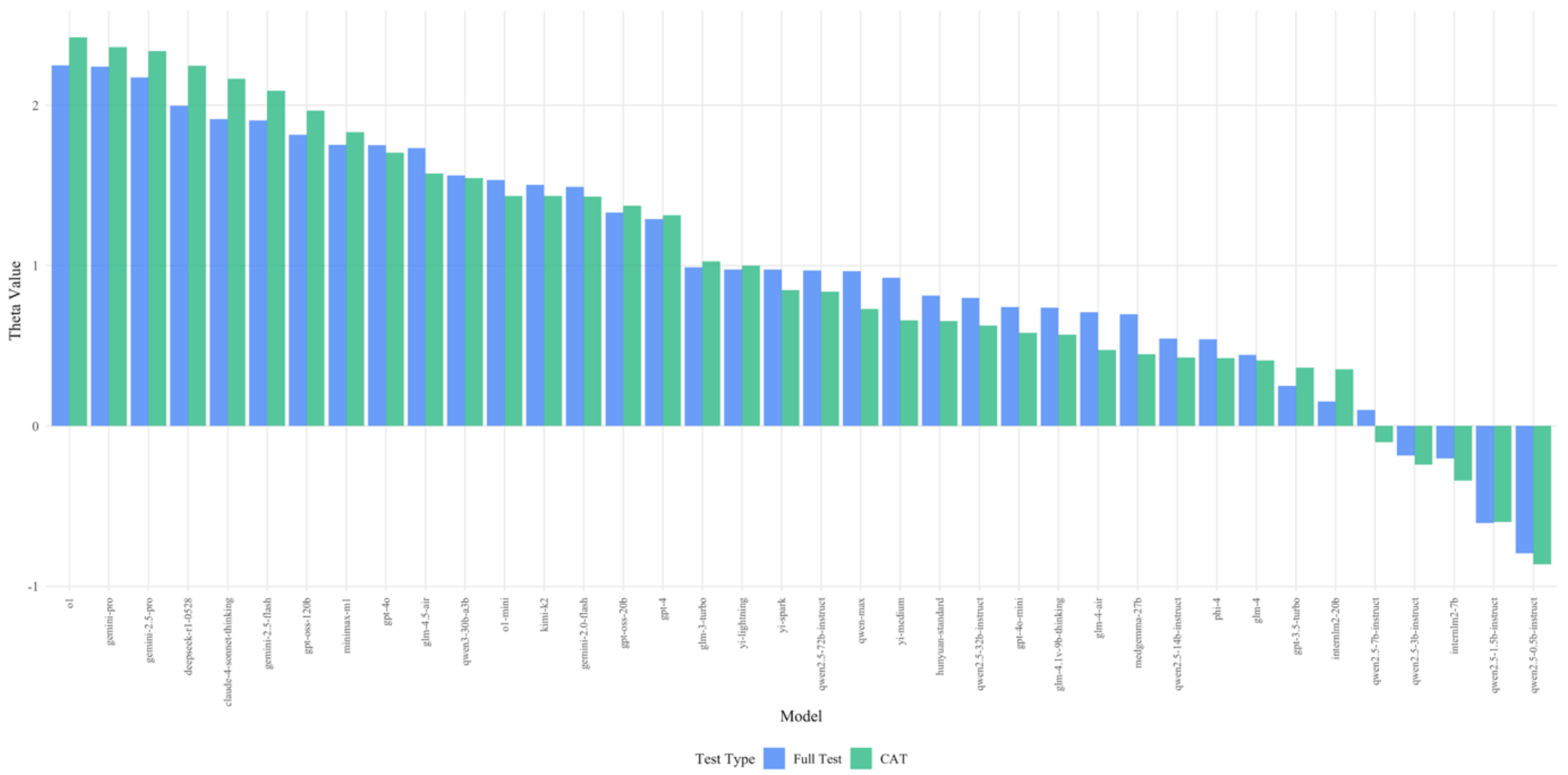

**Figure 4.** Comparison of LLM Ability Estimates from Full-Bank and CAT Assessments

To rigorously isolate the efficiency gains attributable to the adaptive algorithm, we compared the CAT results against a RS baseline. To ensure a robust comparison and minimize stochastic noise, the RS baseline was calculated by averaging $\theta$ estimates over 100 independent simulation runs for each model, using the same test length as CAT (see Table 2). As illustrated in Figure 5, although the RS baseline achieved a high correlation with the full-bank truth, it exhibited a significant systematic bias. Specifically, RS demonstrated a negative bias of -0.130, which is nearly four times larger than that of the CAT protocol (Bias = -0.033). This manifested as regression to the mean, where RS consistently underestimated the capabilities of high-performing models (e.g., the red triangles for $\theta > 2$ fall notably below the identity line). Furthermore, despite the advantage of averaging 100 runs, the RS baseline still yielded a higher RMSE (0.174) compared to the single-run CAT (RMSE = 0.150). Most critically, RS sampling failed to preserve the ranking of state-of-the-art models. As highlighted in the figure, the true top-performing model (o1) was incorrectly ranked as 3rd by the RS baseline, falling behind Gemini-pro and Gemini-2.5-pro due to the lack of challenging items.

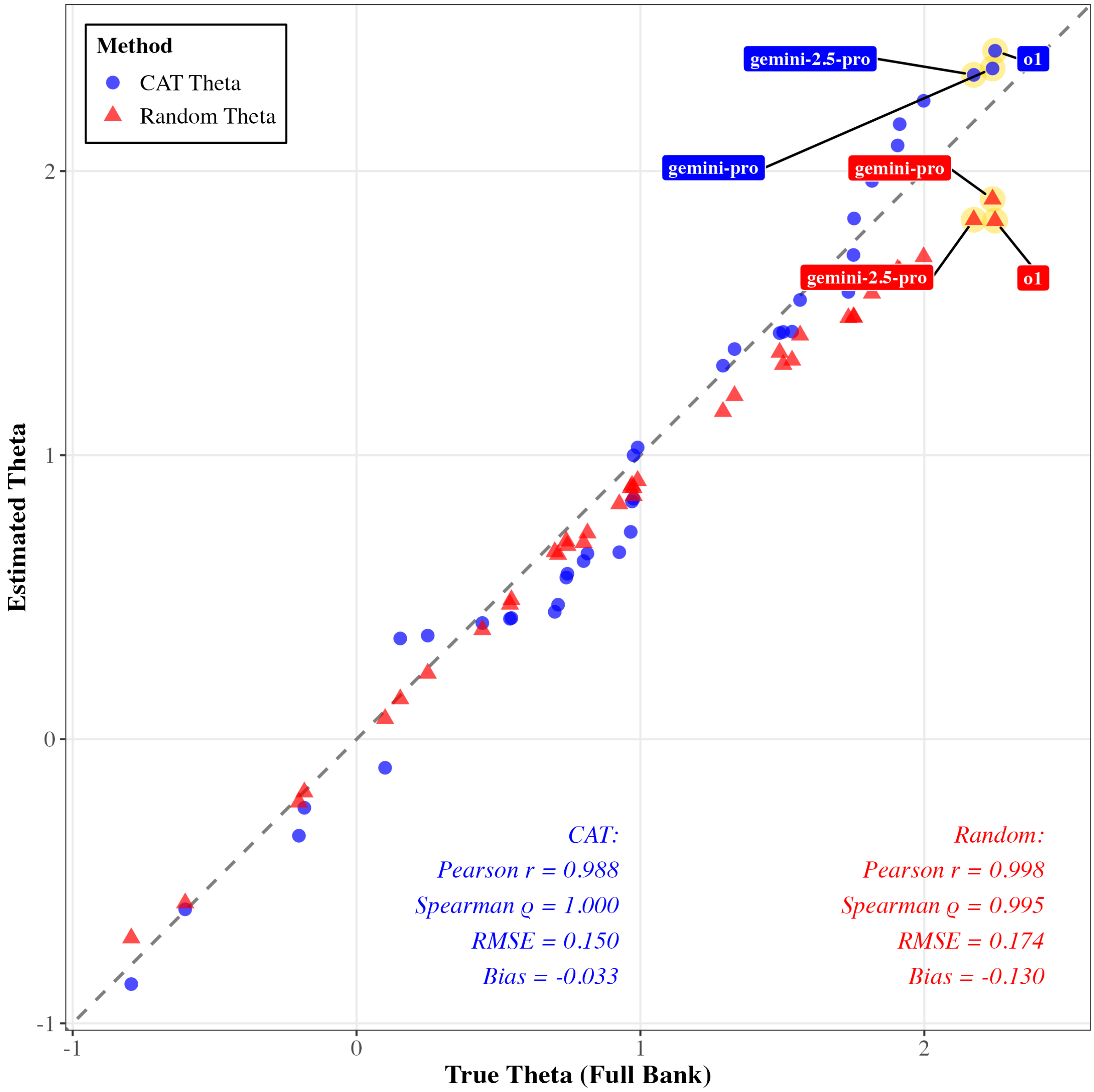

**Figure 5.** Comparison of Ability Estimates ($\theta$) and Methodological Validity

### *3.2.2 Quantification of Cost-Effectiveness Gains*

Having established the exceptional fidelity of the CAT assessment, we next quantified the cost-effectiveness gains it delivers. Figure 6 collectively presents the average reduction in test length, token consumption, and time consumption achieved by our CAT protocol compared to the full-bank benchmark.

The results reveal that the foundational cost-effectiveness gain stems from a drastic reduction in the number of required items. As shown in Figure 6a, the CAT protocol required an average of only 37 items to meet the high-reliability stopping criterion. Compared to the 2,815 items in the full bank, this represents an average test length reduction of 98.7%.

This dramatic decrease in test length translated directly into massive savings in computational resources, which are critical for practical LLM evaluation. The average token consumption decreased from over 1.77 million tokens for the full test to just approximately 0.03 million for the CAT, resulting in an Average Token Reduction of 98.3% (Figure 6b). Similarly, the average time required for an evaluation was reduced from 24,673 seconds (approximately 6.85 hours) to a mere 505 seconds (approximately 8.4 minutes), yielding an Average Time Reduction of 98.0% (Figure 6c).

These findings empirically confirm that our proposed CAT strategy alters the economics of LLM evaluation. It enables a level of measurement fidelity previously achievable only through exhaustive, cost-prohibitive methods, but at a mere 1.3% of the item cost and 2% of the computational and temporal cost.

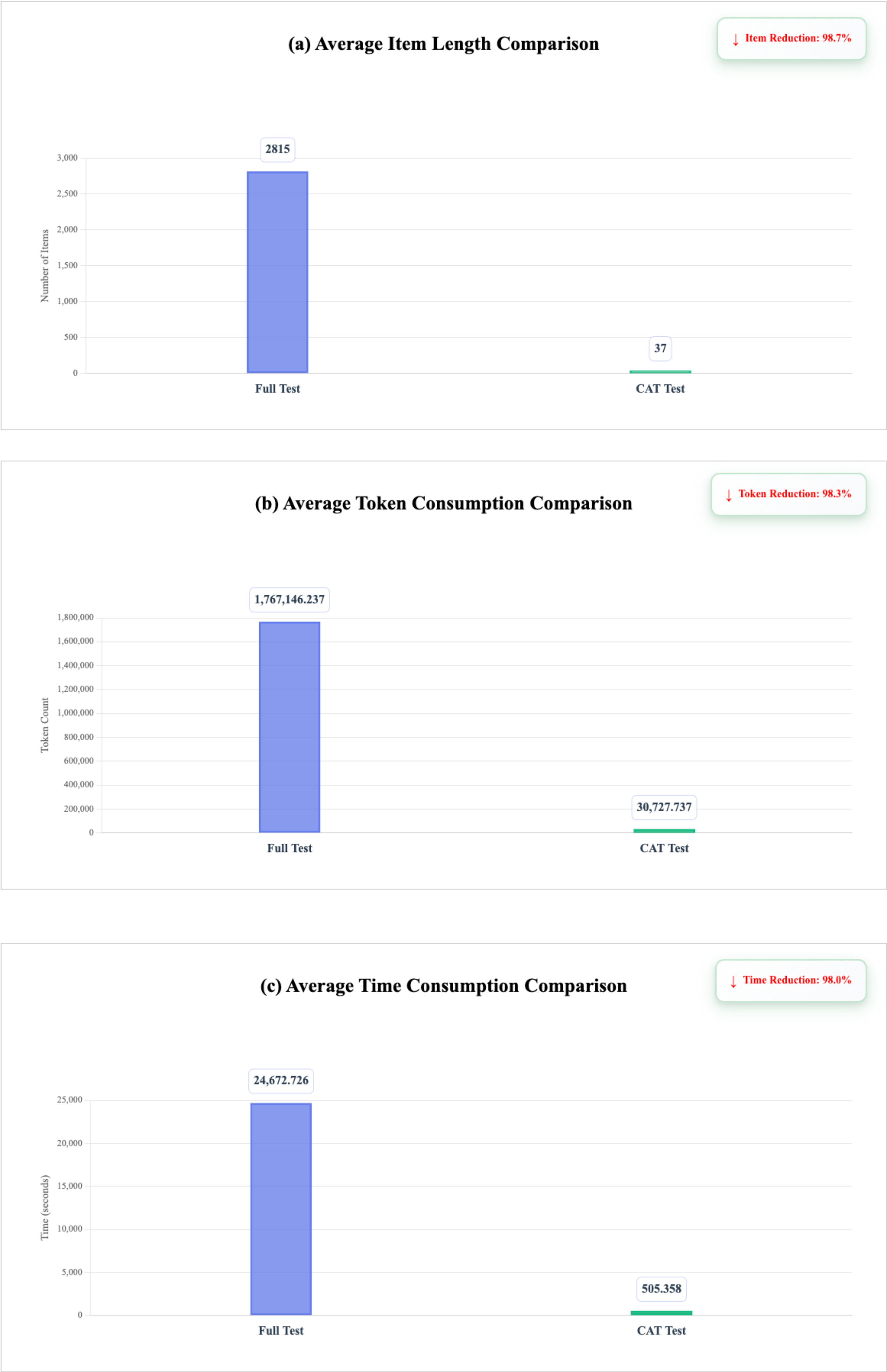

**Figure 6.** Cost-Effectiveness Gains Achieved by the CAT Protocol

## 4. Discussion

This study shows that CAT can be operationalized to evaluate LLMs on a nationally calibrated, secure medical item bank with substantial efficiency gains while preserving comparative fidelity. Under a standardized protocol, our optimal CAT reduced evaluation cost by approximately 98% and preserved the full-bank rank-ordering in our evaluated set. These results support the use of psychometrically grounded adaptive evaluation as a scalable measurement layer for standardized medical knowledge benchmarking.

A practical implication is that adaptivity can markedly reduce the marginal cost of benchmarking at scale. Using empirically observed testing characteristics (e.g., an average of 37 items per administration and per-item token usage) and projecting them to a large benchmark scenario (Med-HALT; 59,254 items[10]), we estimate that the cost of a full-length evaluation would be on the order of $1,475 under the stated pricing assumptions, whereas the CAT-based evaluation would be under $5. Because absolute costs depend on prompting details and pricing, this calculation is presented as an illustrative example rather than a fixed claim. Nevertheless, the orders-of-magnitude reduction suggests that more frequent, protocol-controlled model monitoring (e.g., across versions or updates) becomes feasible compared with exhaustive full-bank testing, enabling evaluation evidence to be accumulated longitudinally within model governance workflows.

At the same time, the scope of inference should be interpreted conservatively. Our evaluation relies on MCQs, which primarily operationalize standardized medical knowledge in an exam-like format and may not fully capture real-world clinical reasoning. Recent studies caution that medical MCQ performance can be inflated by test-taking heuristics and may not reliably indicate clinically meaningful understanding or readiness[34, 35]. While we agree that this is a critical validity concern for general benchmarks, our supplementary cross-modality analysis (Section 3.2.1) mitigates this risk within our specific framework. We observed a strong correlation ($r \approx 0.83$) between our CAT estimates and performance on open-ended generative tasks. This indicates that, for this rigorously calibrated item bank, MCQ scores serve as a robust proxy for genuine medical knowledge rather than mere artifacts of test-taking heuristics. At the same time, consistent with this literature, we do not interpret CAT-based $\theta$ as clinical ability; instead, $\theta$ is an operational proficiency index on a calibrated, secure knowledge bank under a fixed evaluation protocol[35]. This positioning also aligns with observed real-world deployments of clinical LLM assistants in electronic health records contexts, where primary use cases emphasize summarization, information retrieval, and documentation support rather than autonomous clinical decision-making[36].

Furthermore, LLMs are not human examinees and do not imply a single psychological latent trait. In this study, IRT is used operationally. The human calibration sample defines the $\theta$ scale and item parameters under standard assumptions (supported by essential unidimensionality and minimal local dependence), and LLM's $\theta$ is interpreted as a standardized proficiency index on this calibrated bank rather than a cognitive construct. We further minimized context-induced dependencies (independent items, temperature = 0), and the concordance between CAT-based and full-bank ordering provides practical evidence that this scoring yields a consistent comparative mapping within this paradigm.

Several limitations motivate future work. First, our findings are anchored in a single nationally calibrated item bank, generalizability may differ under other blueprints such as the USMLE[18] or European systems. For example, USMLE Step 1 typically emphasizes foundational biomedical mechanisms (e.g., pathophysiology/pharmacology) whereas Step 2 CK places greater weight on clinical diagnosis and management in vignette-based “single best answer” formats, which can shift item-parameter distributions and information functions. Therefore, CAT stopping rules and content constraints should be re-tuned under each exam blueprint, and cross-bank comparisons require psychometric linking/equating rather than assuming transferability of thresholds. Future work should replicate the framework on USMLE- and European-aligned banks and, where needed, apply psychometric linking/equating to support cross-bank comparability. Second, LLMs can exhibit non-monotonic failure patterns (e.g., succeeding on complex items while missing basic safety-critical knowledge). Recent analyses have emphasized the persistent gap between strong benchmark results and the scarcity of rigorous clinical trials for medical AI, underscoring the risk of overreliance on offline evaluations when making deployment decisions[37]. Therefore, we emphasize that the current CAT framework is primarily an efficiency-oriented instrument for quantifying knowledge proficiency, rather than a safety-oriented clearance tool. To address this risk, future adaptive platforms should incorporate safety-aware constraints, such as a mandatory core “safety-and-fundamentals” gate with hard-stop items, and should extend adaptive measurement beyond MCQs toward workflow-aligned task formats with auditable scoring protocols. Until such systems are matured, this adaptive benchmarking must be complemented by dedicated safety evaluations (e.g., adversarial testing or foundational checklists) prior to any real-world deployment.

Overall, our results establish a practical and cost-efficient pathway for adaptive, psychometrically grounded benchmarking of standardized medical knowledge in LLMs, while clarifying the boundaries of what such scores can and cannot support in safety-critical contexts.

**Code Availability** The program code for the computerized adaptive testing system developed in this study will be made publicly available on GitHub upon publication. [https://github.com/zjiang4/LLM-CAT]

# Supplementary Materials

**Appendix A.** Technical Details of the Ability Estimation Method.

**Appendix B.** Technical Details of the Maximum Fisher Information Item Selection Method.

**Appendix C.** Evaluated Models and Their Specifications.

**Appendix D.** Example of Prompt, Question, and LLM Response.

**Appendix E.** Supplementary Analysis on Construct Validity via Cross-Modality Evaluation.

**Appendix A.** Technical Details of the Ability Estimation Method

In a Computerized Adaptive Testing system, the estimation of an examinee's latent ability, denoted by $\theta$, is a core iterative step. To ensure robust and stable ability estimates throughout the testing process, our study employed the Expected A Posteriori (EAP) estimation method. EAP is a Bayesian approach that is widely adopted in operational CAT programs due to its superior performance with short tests. The fundamental idea of EAP is to estimate an examinee's ability by calculating the mean of the posterior distribution of $\theta$, given their vector of item responses. Formally, given an examinee's response vector $\boldsymbol{X} = (x_1, x_2, \cdots, x_j)$ to $j$ items, the EAP estimate of their ability, $\hat{\theta}_{EAP}$, is calculated as the expected value of $\theta$ with respect to its posterior distribution:

$$\hat{\theta}_{EAP} = \boldsymbol{E}(\theta|\boldsymbol{X}) = \frac{\int_{-\infty}^{\infty} \theta L(\boldsymbol{X}|\theta) g(\theta) d\theta}{\int_{-\infty}^{\infty} L(\boldsymbol{X}|\theta) g(\theta) d\theta}$$

where: $L(\boldsymbol{X}|\theta)$ is the likelihood function of observing the response pattern $\boldsymbol{X}$ given an ability level of $\theta$. Under the assumption of local independence in IRT, this is the product of the probabilities of the individual item responses: $L(\boldsymbol{X}|\theta) = \prod_{j=1}^{J} P_j(x_j|\theta)$. $g(\theta)$ is the prior distribution of ability in the population. In CAT applications, this is typically assumed to be a standard normal distribution, $N(0,1)$, reflecting the distribution of the reference population upon which the item bank was calibrated. The numerator of the EAP formula is the integral of the posterior distribution weighted by $\theta$,, while the denominator corresponds to the marginal likelihood—the integral of the likelihood weighted by the prior—which serves as a normalizing constant. By incorporating information from the prior distribution, EAP provides a stable and finite ability estimate even under extreme response patterns.

**Appendix B.** Technical Details of the Maximum Fisher Information Item Selection Method

In a Computerized Adaptive Testing (CAT) framework, the item selection strategy plays a crucial role in ensuring efficient measurement of examinee ability. Among the most widely used methods is the Maximum Fisher Information (MFI) criterion, which selects the next item based on its information contribution at the current ability estimate.

Formally, under the Item Response Theory (IRT) model, the Fisher information of item $j$ at ability level $\theta$ is defined as:

$$I_j(\theta) = \mathrm{Var}\left(\frac{\partial}{\partial\theta}\ln P\left(x_j|\theta\right)\right),$$

where $P(x_j|\theta)$is the probability of a particular response to item j at ability level $\theta$. For dichotomous models such as the 2-parameter logistic (2PL) model,

$$P_j(\theta_i) = \frac{exp[\alpha_j(\theta_i - \beta_j)]}{1 + exp[\alpha_j(\theta_i - \beta_j)]},$$

where $\alpha_j$is the discrimination parameter and $\beta_j$is the difficulty parameter. The Fisher information simplifies to:

$$I_j(\theta) = \alpha_j^2\, P_j(\theta)\left[1 - P_j(\theta)\right].$$

At each stage of CAT, given the current provisional ability estimate $\hat{\theta}$ (obtained, for example, via EAP or MLE), the algorithm computes $I_j(\hat{\theta})$ for all eligible items in the remaining pool. The next item is then chosen as the one that maximizes this information:

$$j^* = \arg\max_{j\in\mathcal{J}} I_j\left(\hat{\theta}\right),$$

where $\mathcal{J}$ denotes the set of items not yet administered.

**Appendix C.** Evaluated Models and Their Specifications

| Model Name | Type | Category | Access Method |
|---|---|---|---|
| claude-4-sonnet-thinking | Commercial | Reasoning | API |
| deepseek-r1-0528 | Open-Source | Reasoning | API |
| gemini-2.0-flash | Commercial | Non-Reasoning | API |
| gemini-2.5-flash | Commercial | Reasoning | API |
| gemini-2.5-pro | Commercial | Reasoning | API |
| gemini-pro | Commercial | Reasoning | API |
| glm-3-turbo | Commercial | Non-Reasoning | API |
| glm-4-air | Open-Source | Non-Reasoning | API |
| glm-4.1v-9b-thinking | Open-Source | Reasoning | API |
| glm-4.5-air | Open-Source | Reasoning | API |
| glm-4 | Open-Source | Non-Reasoning | API |
| gpt-3.5-turbo | Commercial | Non-Reasoning | API |
| gpt-4 | Commercial | Non-Reasoning | API |
| gpt-4o-mini | Commercial | Non-Reasoning | API |
| gpt-4o | Commercial | Non-Reasoning | API |
| gpt-oss-120b | Open-Source | Reasoning | API |
| gpt-oss-20b | Open-Source | Reasoning | API |
| hunyuan-standard | Commercial | Non-Reasoning | API |
| internlm2-20b | Open-Source | Non-Reasoning | Local Deployment |
| internlm2-7b | Open-Source | Non-Reasoning | Local Deployment |
| kimi-k2 | Open-Source | Non-Reasoning | API |
| medgemma-27b | Open-Source | Non-Reasoning | Local Deployment |
| minimax-m1 | Open-Source | Reasoning | API |
| o1-mini | Commercial | Reasoning | API |
| o1 | Commercial | Reasoning | API |
| phi-4 | Open-Source | Non-Reasoning | API |
| qwen-max | Commercial | Non-Reasoning | API |
| qwen2.5-0.5b-instruct | Open-Source | Non-Reasoning | API |
| qwen2.5-1.5b-instruct | Open-Source | Non-Reasoning | API |
| qwen2.5-14b-instruct | Open-Source | Non-Reasoning | API |
| qwen2.5-32b-instruct | Open-Source | Non-Reasoning | API |
| qwen2.5-3b-instruct | Open-Source | Non-Reasoning | API |
| qwen2.5-72b-instruct | Open-Source | Non-Reasoning | API |
| qwen2.5-7b-instruct | Open-Source | Non-Reasoning | API |
| qwen3-30b-a3b | Open-Source | Reasoning | API |
| yi-lightning | Commercial | Non-Reasoning | API |
| yi-medium | Commercial | Non-Reasoning | API |
| yi-spark | Commercial | Non-Reasoning | API |

**Appendix D.** Example of Prompt, Question, and LLM Response

**Prompt:**
## Please answer the following single-choice question.
## Instructions: 1. Choose the correct option letter from the given choices; 2. Output ONLY the letter of the correct answer. Do not include any explanations or extra text.
## Example:
Question:
This is an example question stem where you need to choose the correct answer.
Options:
A.Incorrect answer
B.Incorrect answer
C.Correct answer
D.Incorrect answer
E.Incorrect answer
Answer: C
## Now, here is the question you need to answer. Again, output ONLY the letter of the correct answer:
{questions}

**The Example of Question:**
Question:
A 40-year-old male from Brazil visits his primary care physician after experiencing persistent oral thrush, for which he is prescribed fluconazole. One month later, he reports a dry cough, unexplained night sweats, and breathlessness during physical activity. A chest CT performed upon his hospital admission shows diffuse bilateral ground-glass opacities. Which of the following is the most likely diagnosis?
Options:
A. Bacterial pneumonia
B. Late onset asthma
C. Tuberculosis (TB)
D. Immunosuppression-related pneumonia
E. Pneumocystis jirovecii pneumonia (PCP)

**The Example of LLM Response:**
E

**Appendix E.** Supplementary Analysis on Construct Validity via Cross-Modality Evaluation

### E.1 Rationale and Objective

A key challenge in medical benchmarking is ensuring that performance on multiple-choice questions (MCQs) reflects genuine medical knowledge rather than test-taking heuristics or pattern recognition. To address concerns regarding the construct validity of our CAT framework, we conducted a cross-modality validation study.

The objective of this supplementary analysis was to evaluate the correlation between our CAT-based proficiency estimates (MCQ format) and model performance on a generative, open-ended medical QA benchmark and assess the robustness of this correlation using a Dual-Judge Protocol involving both a state-of-the-art generalist LLM and a specialized medical LLM to mitigate potential evaluator bias.

### E.2 Methodology and Dataset

We utilized the "Medical Knowledge" subset of the LLMEval-Med benchmark[1], comprising 141 high-quality items designed to assess standardized medical knowledge in an open-ended format. Each data entry includes a question, a reference answer authored and refined by medical experts, a prompt simulating real-world usage, and a checklist specifying key content requirements.

Unlike the MCQ task format, the models in this validation phase were presented with the question stem only and were required to generate the answer directly. This format removes the possibility of guessing based on option elimination strategies. Due to the rapid lifecycle of LLMs, 4 of the original 38 models were deprecated or became inaccessible via API during the revision period (February 2026). Consequently, this analysis was performed on the remaining 34 accessible models. This subset retains the diversity of the original cohort, spanning from small open-source models to large proprietary frontiers.

---

[1] Zhang M, Shen Y, Li Z, et al. LLMEval-Med: A real-world clinical benchmark for medical large language models with physician validation. Findings Assoc Comput Linguist EMNLP. 2025:4888–4914.

To mitigate subjective variability in scoring open-ended responses, we adopted the "LLM-as-a-Judge" methodology, employing State-of-the-Art (SOTA) LLMs as automated evaluators. The evaluation includes four core components: Prompt (defining the role and task background), Question, Response, and Reference Answer. During scoring, the judge model utilizes structured scoring prompts to ensure consistent evaluation criteria. These prompts explicitly define scoring dimensions, metrics, and rules, thereby reducing subjective variability. This structured approach minimizes the hallucination or bias often associated with unstructured LLM grading. For detailed item content, prompt designs, and specific scoring examples, please refer to the Appendix of the original LLMEval-Med publication.

**E.3 Dual-Judge Evaluation Protocol**

To ensure the robustness of our scoring mechanism and mitigate potential evaluator bias, we implemented a Dual-Judge Protocol utilizing two distinct large language models with complementary strengths. First, we employed GPT-5.2 as a generalist evaluator. Second, to address domain-specific nuances within the Chinese medical context, we utilized Baichuan-M3 as a specialist evaluator, which is built for clinical decision support. The temperature parameter was fixed at 0.0 to ensure deterministic outputs. The top-*p* value was set to 1.0 to allow sampling from the full probability distribution without truncation.

## E.4 Results

We calculated the Pearson correlation coefficient between the CAT-estimated Theta from the main study and the aggregate accuracy scores from the generative task. As illustrated in Supplementary Figure E1, the cross-modality analysis revealed a robust linear relationship between the CAT-estimated proficiency ($\theta$) and the open-ended generative performance. Notably, the Pearson correlation coefficient remained high and consistent across both evaluators, with GPT-5.2 yielding an $r$ of 0.829 ($p < 0.001$) and Baichuan-M3 similarly yielding an $r$ of 0.831($p < 0.001$) .

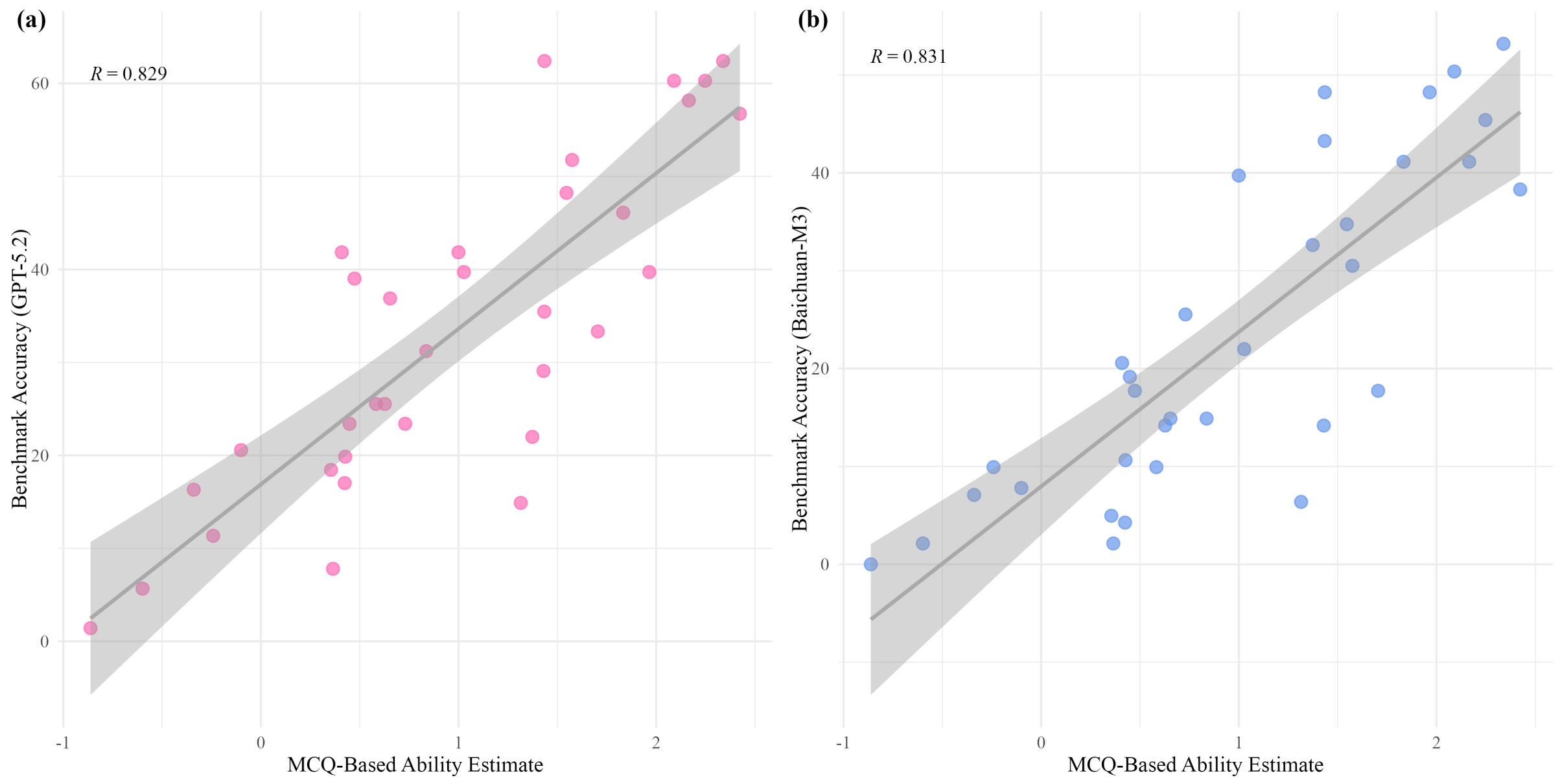

**Figure E1**. Cross-Modality Validation of Construct Validity